\documentclass[final]{cvpr}

\usepackage{times}
\usepackage{epsfig}
\usepackage{graphicx}
\usepackage{amsmath}
\usepackage{amssymb}

\usepackage{caption}
\usepackage{subcaption}

\usepackage[pagebackref=true,breaklinks=true,colorlinks,bookmarks=false]{hyperref}

\def\cvprPaperID{2931} 
\def\confYear{CVPR 2021}

\begin{document}

\title{DETR for Crowd Pedestrian Detection}

\author{
Matthieu Lin $^1$
\and
Chuming Li $^2$
\and
Xingyuan Bu $^2$
\and
Ming Sun $^2$
\and
Chen Lin $^3$
\and
Junjie Yan $^2$
\and
Wanli Ouyang  $^4$
\and
Zhidong Deng  $^1$ 


\and
$^1$Department of Computer Science and Technology, Tsinghua University.\thanks{State Key Laboratory of Intelligent Technology and Systems, Beijing National Research Center for Information Science and Technology,
and Center for Intelligent Connected Vehicles and Transportation.}
 \\
$^2$SenseTime Group Limited, $^3$University of Oxford , $^4$The University of Sydney 
\and
\texttt{\{lin-yh19@mails, michael@\}.tsinghua.edu.cn} \\
\texttt{\{lichuming, sunming1, yanjunjie \}@sensetime.com} \\
\texttt{sefira32@gmail.com},  \texttt{chen.lin@eng.ox.ac.uk}, \texttt{wanli.ouyang@sydney.edu.au}

}






\maketitle

\begin{abstract}

	Pedestrian detection in crowd scenes poses a challenging problem due to the heuristic defined mapping from anchors to pedestrians and the conflict between NMS and highly overlapped pedestrians. The recently proposed end-to-end detectors(ED), DETR and deformable DETR, replace hand designed components such as NMS and anchors using the transformer architecture, which gets rid of duplicate predictions by computing all pairwise interactions between queries. Inspired by these works, we explore their performance on crowd pedestrian detection. Surprisingly, compared to Faster-RCNN with FPN, the results are opposite to those obtained on COCO. Furthermore, the bipartite match of ED harms the training efficiency due to the large ground truth number in crowd scenes. In this work, we identify the underlying motives driving ED's poor performance and propose a new decoder to address them. Moreover, we design a mechanism to leverage the less occluded visible parts of pedestrian specifically for ED, and achieve further improvements. A faster bipartite match algorithm is also introduced to make ED training on crowd dataset more practical. The proposed detector PED(Pedestrian End-to-end Detector) outperforms both previous EDs and the baseline Faster-RCNN on CityPersons and CrowdHuman. It also achieves comparable performance with state-of-the-art pedestrian detection methods. Code will be available on  \url{https://github.com/Hatmm/PED-DETR-for-Pedestrian-Detection}.

\end{abstract}

\section{Introduction}
\label{introduction}
Pedestrian detection is a critical research field due to its wide application in self-driving, surveillance and robotics. In recent years, promising improvements of pedestrian detection have been made. However, pedestrians in occluded or crowd scenes remain difficult to detect accurately.

Pedestrian detection involves two fundamental challenges that remain to be addressed: (1) mapping features to instances and (2) duplicate prediction removal. 
In the former case (1), most detectors build a heuristic mapping from points on the convolution neural network (CNN) feature maps to ground truth (GT) bounding box. A point is assigned to a GT if it has a small distance to the GT's center or the anchor defined on it has a high intersection over union(IOU) with the GT. 
\begin{figure}
	\begin{center}
		
		\includegraphics[width=0.8\linewidth]{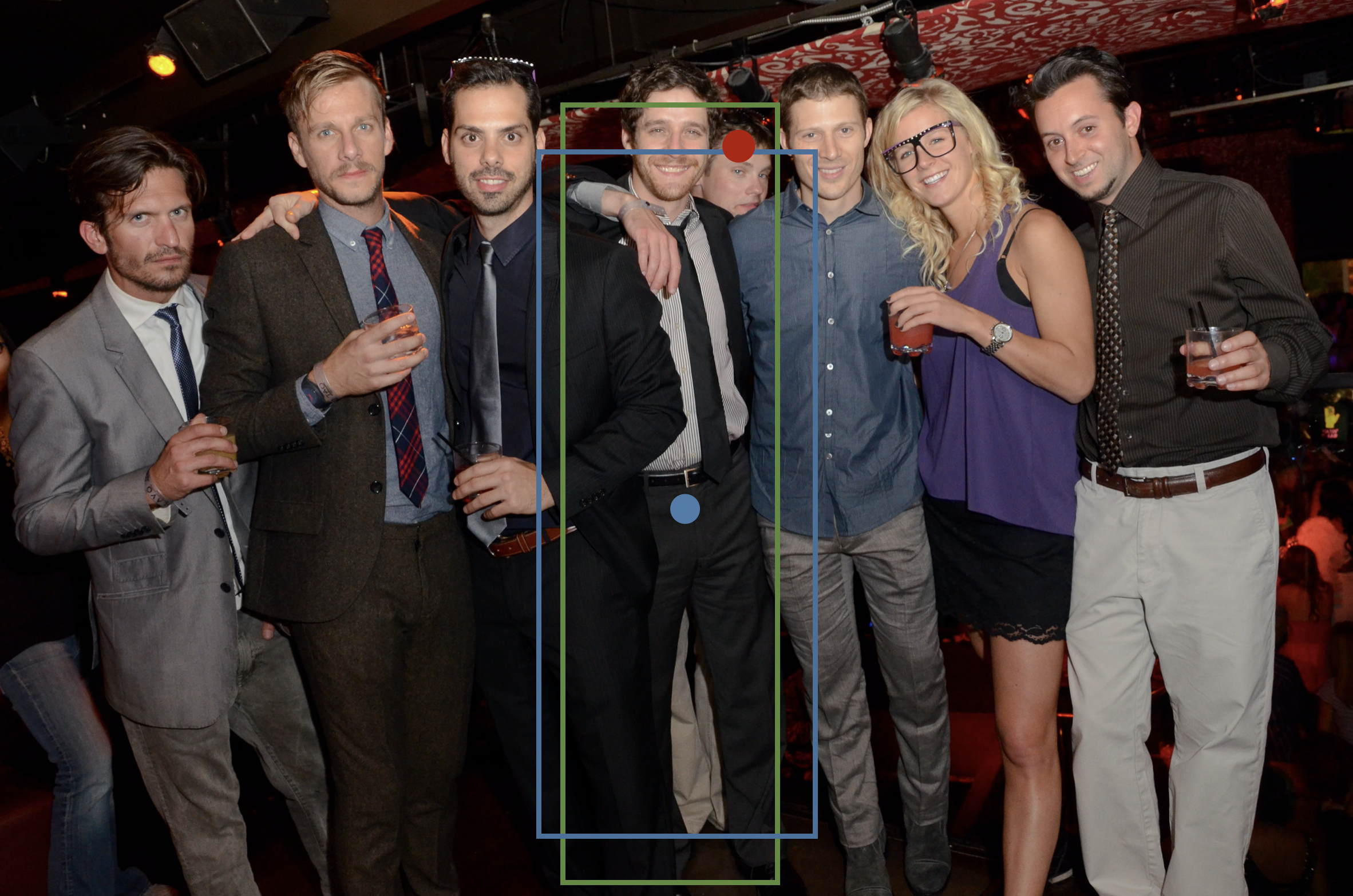}%
		
	\end{center}
	\caption{The green box is the visible region of the person behind. The blue box and dot are the anchor and point matching the person. As we can observe, the point or point of the anchor is lying on the body of the front person, hence making the mapping ambiguous. While deformable DETR is able to adaptively refine its attention position to the features of the visible part (red point)}
	\label{fig:anchor}
\end{figure}
Nevertheless, due to the highly overlapped pedestrians with appearance variance. Points lying in the central part of one GT is likely to be mapped to another one. It means the heuristically defined mapping is ambiguous.
In Figure \ref{fig:anchor}, we show that both distance-based mapping and anchor-based mapping suffer from ambiguity even with visible annotations.
In the latter case (2), duplicate proposals of a single GT is usually provided by modern detectors and need a post-process mechanism to filter out. However, the widely used non-maximum suppression(NMS) relies on intersection-over-union(IOU) and fails in crowd scenes. Because duplicate proposals and another GT's proposals may both have high IOUs with a GT's true positive proposal.

Existing solutions to pedestrian detection predominantly focus on two types of improvements. Some researchers explore using more distinctive body parts, \eg, head or visible region, 
\begin{figure*}
	
	\begin{center}
		
		\includegraphics[width=0.8\linewidth]{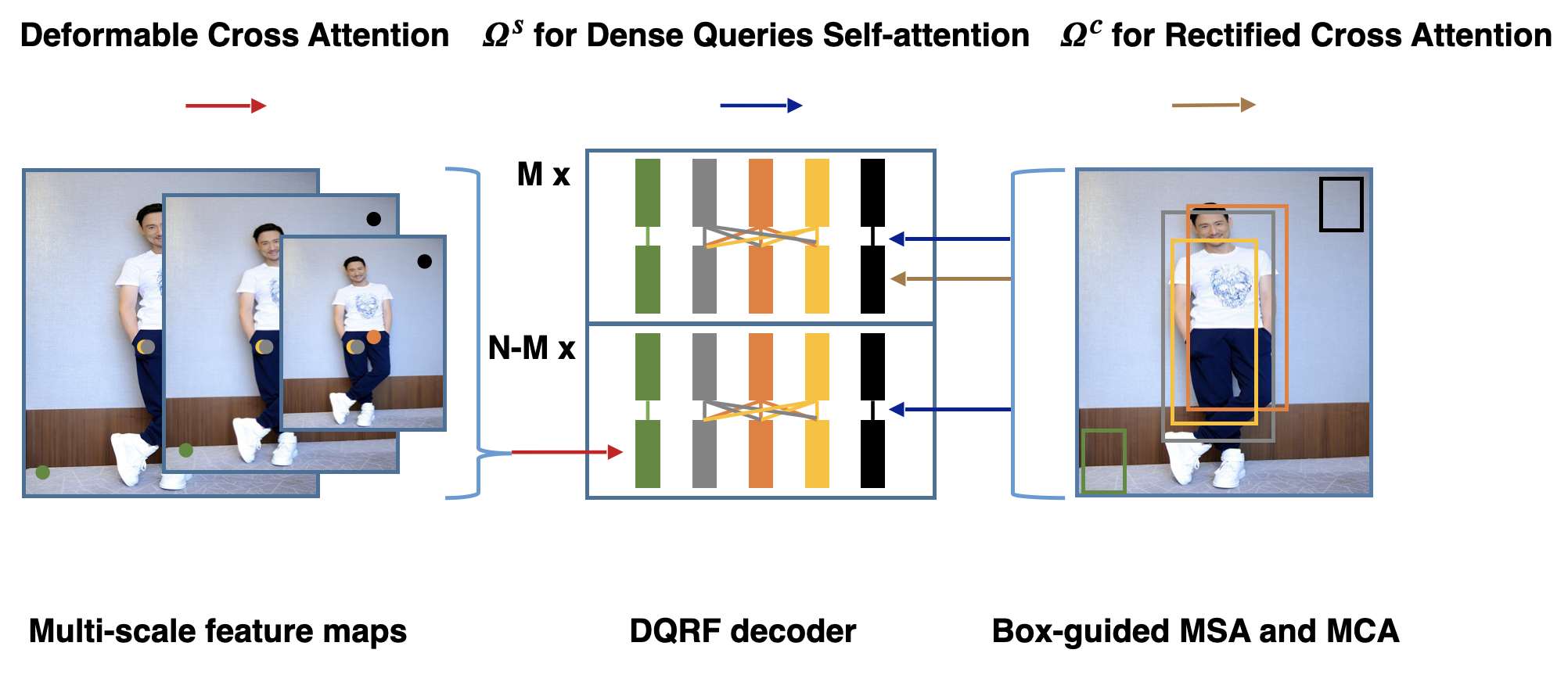}%
		
	\end{center}
	
	\caption{Overview of our DQRF decoder with $N$ DQ layers and $M$ RF layers. The encoder architecture is kept as in~\cite{zhu2020deformable}.}
	
	\label{fig:dqr_architecture}
\end{figure*}
and use it to learn extra supervision, re-weight feature maps or guide the anchor selection~\cite{zhang2018occlusion,pang2019mask,chi2020relational}. Some other works propose clever methods to introduce more signals to make duplicate proposals and close GT's proposals more distinguishable, including neighbor GT's existence and direction, IOU between visible regions and local density~\cite{liu2019adaptive,huang2020nms,chi2020relational}. Both the two types of works achieve significant improvements and partly solve the two challenges mentioned above.

The recently proposed end-to-end detectors, DETR and deformable DETR~\cite{carion2020end,zhu2020deformable}, perform comparably or even better on common objects detection. In the subsequent part, we use 'DETR' to refer to DETR and deformable DETR.
We identify DETRs' two properties which imply their natural advantages on pedestrian detection over the former works. On one hand, DETR is query-based and does not rely on heuristic design of the mapping between feature map points and GTs. Instead, DETR queries adaptively determine their effective attention areas feature maps and the corresponding objects. On the other hand, DETR learns a bipartite match between queries and GTs. The match assigns an individual query for each GT, without any duplicate proposal. The two properties suggest DETR's potential in solving the two challenges of pedestrian detection specifically.

\begin{table}[!htbp]
	
	\begin{center}
		\begin{tabular}{l|c|c|c|c}
			\hline
			Model & Epochs & GPU days &AP & MR$^{-2}$\\
			\hline
			Faster-RCNN & 20 &   0.75 &  85.0 & 50.4 \\
			DETR   & 300 & 223.7 & 66.12 & 80.62  \\
			+Deformable  & 50 &  8.4 &  86.74 & 53.98  \\
			\hline
		\end{tabular}
	\end{center}
	\caption{Comparison of DETRs and Faster-RCNN + FPN on CrowdHuman. All models are trained on 8 Tesla V-100s. For both DETRs we increase the number of object queries to 400 due to the large GT number of CrowdHuman.}
	\label{tab:fasterrcnnvsdetr}
\end{table}

We explore DETR's performance on pedestrian detection and compare them with Faster-RCNN~\cite{ren2015faster}, which is the standard baseline used in both DETR works and pedestrian detection works. Unfortunately, the results are opposite of that on COCO~\cite{lin2014microsoft}. Table~\ref{tab:fasterrcnnvsdetr} indicates that both original DETR and deformable DETR perform much worse than Faster-RCNN on CrowdHuman~\cite{shao2018crowdhuman}. Additionally, the training of DETR on dataset of crowd pedestrians is quite time-consuming due to the standard KM algorithm~\cite{km} used for bipartite match that has a time complexity cubic of the GT number. The bipartite match costs about 2 times the time of the detectors' forward plus backpropagation on CrowdHuman, thus bottlenecks DETR's training.

We analyse the reasons behind deformable DETR's poor performance on pedestrian detection due to its advantages over original DETR. We find that its (1) sparse uniform queries and (2) weak attention field harm the performance. The decoder of deformable DETR learns a mapping from sparse uniformly distributed queries to a naturally local dense pedestrian cluster. As we discuss in Section~\ref{subsec:dq}, such mapping is ambiguous and results in missed GTs. The attention positions of the decoder over the feature maps is also problematic. The attention positions is adaptively learnt during training, while they do not converge to a rectified and compact position set which covers the corresponding GT well. On the contrary, they tend to cover more than one GTs or not extensible enough for large persons.
From these observations, we propose a decoder with \textbf{d}ense \textbf{q}ueries and \textbf{r}ectified attention \textbf{f}ield (DQRF). DQRF significantly improves DETR on pedestrian detection and closes the gap between DETR and Faster-RCNN.
Furthermore, we explore how to leverage annotations of visible regions and establish a visible region based set supervision, namely V-Match, together with a data augmentation which is visible region aware, which enhance DETR's performance further.
Finally, we design a heuristic improvement of KM algorithm based on the prior that GT tends to match its close proposals in the bipartie matching of ED. The resulted Fast-KM gains up to 10x speed-up, making DETR practical on pedestrian detection tasks. 

Our contributions are summarized as:
\begin{itemize}
	\item We conduct in-depth analysis of DETR for pedestrian detection task and identify the problems when directly applying DETR for pedestrian detection.
	
	\item We propose a new decoder for DETR, DQRF, which significantly improves DETR on pedestrian detection and closes the gap between DETR and Faster-RCNN. A Fast-KM is also proposed to make DETR practical on pedestrian detection.
	\item We further explore the leverage of annotations of visible regions specifically for DETR and establish a visible region based set supervision, V-Match, together with a data augmentation which is aware of visible region.
\end{itemize}
The resulted \textbf{PE}destrian-specific \textbf{D}ETR, namely PED, outperforms competitive Faster-RCNN and achieves comparable performance with state-of-the-art results on the challenging CrowdHuman and CityPersons~\cite{zhang2017citypersons} benchmarks. We hope that our work can serve as a new baseline for end-to-end detectors on pedestrian detection in the same way DETR and deformable DETR for common object detection.


\section{Related Work}
\label{relatedwork}
\noindent \textbf{Generic Object Detection.}
In the era of deep learning, most generic object detection could be roughly divided into two categories, \ie, two-stage detection~\cite{girshick2014rich,he2015spatial,girshick2015fast,ren2015faster} and one-stage detection~\cite{liu2016ssd,redmon2016you,lin2017focal,li2018gradient}, depending on whether an explicit region proposal and pooling process are employed. To further improve the performance, FPN~\cite{lin2017feature,ghiasi2019fpn} and DCN series~\cite{dai2017deformable,zhu2019deformable} are introduced to enhance the feature representation. Meanwhile, iterative prediction~\cite{cai2019cascade} and extra supervision~\cite{he2017mask,jiang2018acquisition,huang2019mask} could yield more precise bounding boxes. Recently, some works try to remove the pre-defined anchor hypothesis over the feature map grid, known as anchor-free detection methods. They tend to use the center~\cite{duan2019centernet} or keypoint~\cite{zhou2019objects,tian2019fcos} instead of anchor box. Relation networks~\cite{hu2018relation} models the relation between different proposals by long-range attention, which is then used to distinguish whether a generated proposal is the unique prediction. However, it still involves hand-crafted rank and box embedding. Different from the above generic object detection methods, DETR~\cite{carion2020end} can generate set prediction direct from the input image.

\noindent \textbf{Pedestrian Detection.}
Pedestrian detection is the fundamental technology for self-driving, surveillance, and robotics. Although great progress has been made in the past decade~\cite{dollar2011pedestrian,zhang2016far,zhang2017citypersons}, pedestrian detection in crowd scenario remains challenging due to the confusing features and dilemmatic NMS~\cite{shao2018crowdhuman}. OR-CNN~\cite{zhang2018occlusion} and MGAN~\cite{pang2019mask} regard the invisible parts as noise and down-weighting those features to deal with the confusing feature. PedHunter~\cite{chi2020pedhunter} applies a stricter overlap strategy to reduce the ambiguity of matching. Using the human head as a clue also be explored in~\cite{chi2020pedhunter,chi2020relational}. For the dilemmatic NMS, adaptive NMS~\cite{liu2019adaptive} first predicts the crowd density, then dynamically adjusts the NMS threshold according to predicted density. The visible box and head box~\cite{huang2020nms,chi2020relational} also increase the performance of the NMS and partly address the essential problem of how to generate an appropriate prediction. Anchor-free methods are also adapted to pedestrian detection in~\cite{liu2019high, xu2020beta}, and they reserves NMS. \cite{stewart2016end} views proposal prediction as a sequence generation, which is an unnecessary attribute for detection. 

\noindent \textbf{DETR.}
Generic object detection methods usually contain a hand-crafted post-process, \ie, NMS. Despite its variant type~\cite{bodla2017soft, he2018softer}, this post-process has no access to the image information and the network feature and must be employed solely. Thus, NMS can not be optimized in an end-to-end manner. The recent DETR~\cite{carion2020end} and deformable DETR~\cite{zhu2020deformable} utilize the encoder-decoder architecture based on the transformer module, which could essentially build context features and remove duplicates. Although they achieve relatively high performance in common detection datasets, they do not work well in pedestrian detection under crowd scenarios, as discussed in the introduction.

\section{Revisit DETR}
\label{preliminary}

\subsection{Decoder} The recent DETRs are based on the architecture of transformer \cite{vaswani2017attention} and assign a unique query for each GT through bipartite matching. Here we briefly formulate the detection process of DETR as follows. 

Let $\boldsymbol{q}$ represents a set of queries with $\boldsymbol{q}_i \in R^C$ and 
$\boldsymbol{x}$ represents the feature map points with $\boldsymbol{x}_i \in R^C$. In the decoder of the transformer, the query set $\boldsymbol{q}$ is iteratively updated via cross-attention between $\boldsymbol{q}$ and $\boldsymbol{x}$ and self-attention among $\boldsymbol{q}$. This process can be formulated as a sequence of functions $F^t$, where $\boldsymbol{q}^t = F^t(\boldsymbol{q}^{t-1},\boldsymbol{x})$, and $t \in {1,...,T}$ with $T$ denoting the number of decoder layers.
We further decompose $F^t$ into $F_c^t$ and $F_s^t$. $F_c^t(\boldsymbol{q},\boldsymbol{x})$ represents the cross attention between $\boldsymbol{q}$ and $\boldsymbol{x}$ and $F_s^t(\boldsymbol{q})$ represents the self attention among $\boldsymbol{q}$. Hence, $F^t(\boldsymbol{q}^{t-1},\boldsymbol{x})=F_c^t(F_s^t(\boldsymbol{q}^{t-1}),\boldsymbol{x})$. 

\subsection{Multi-Head Attention}
The functions $F_c^t$ and $F_s^t$ of DETR are based on transformer's multi-head attention. In both the two DETRs, $F_s^t$ consists of a standard multi-head self attention module(MSA) followed by a multiple linear projection(MLP) as in Eq. \ref{eq:fsmsa}, \ref{eq:fsmlp}, where LN means layer normalization.

\begin{align}
\boldsymbol{q}_{msa}^{t-1} &= LN(MSA(\boldsymbol{q}^{t-1})+\boldsymbol{q}^{t-1}), \label{eq:fsmsa} \\
F_s^t(\boldsymbol{q}^{t-1}) = \boldsymbol{q}_{s}^{t-1} &=  LN(MLP(\boldsymbol{q}_{msa}^{t-1})+\boldsymbol{q}_{msa}^{t-1}).  \label{eq:fsmlp}
\end{align}

The cross attention function $F_c^t$ similarly consists of a multi-head cross attention module(MCA) between $\boldsymbol{q}$ and $\boldsymbol{x}$ followed by a MLP (Eq. \ref{eq:fcmca}, \ref{eq:fcmlp}). 
\begin{align}
\boldsymbol{q}_{mca}^{t-1} &= LN(MCA(\boldsymbol{q}_{s}^{t-1}, \boldsymbol{x})+\boldsymbol{q}_{s}^{t-1}), \label{eq:fcmca} \\
F_c^t(\boldsymbol{q}^{t-1},\boldsymbol{x}) = \boldsymbol{q}^t &= LN(MLP(\boldsymbol{q}_{mca}^{t-1})+\boldsymbol{q}_{mca}^{t-1}).  \label{eq:fcmlp}
\end{align}
Both MSA and MCA can be expressed by a basic multi-head attention(MA) module as in Eq. \ref{eq:ma}, \ref{eq:msa}, \ref{eq:mca}. 
\begin{align}
MA(\boldsymbol{q}_i,\boldsymbol{z}) &= \sum_{m=1}^M \boldsymbol{W}_m \sum_{k \in \Omega_{i}} \boldsymbol{A}_{mik} \boldsymbol{W}_m^V \boldsymbol{z}_k ,
\label{eq:ma}\\
MSA(\boldsymbol{q}_i) &= MA(\boldsymbol{q}_i, \boldsymbol{q}) ,
\label{eq:msa}\\
MCA(\boldsymbol{q}_i, \boldsymbol{x}) &= MA(\boldsymbol{q}_i, \boldsymbol{x}) .
\label{eq:mca}
\end{align}
$\boldsymbol{z}$ is a set of vectors where $\boldsymbol{q}_i$ will make attention on. In MSA, $\boldsymbol{z}$ is $\boldsymbol{q}$ itself and in MCA the feature maps $\boldsymbol{x}$.
For each head $m$, a linear projection $\boldsymbol{W}_m^V \in R^{\frac{C}{M} \times C}$ is operated on $\boldsymbol{z}$ to map it to a new representation. An attention weight $\boldsymbol{A}_{mik}$ is applied to weighted sum the representations on positions in $\Omega_{i}$, which is the attention field of $\boldsymbol{q}_i$. Finally, the weighted summed features of all heads are linearly projected to $R^C$ via $\boldsymbol{W}_m$ and summed.

Original DETR and deformable DETR differs in the designs of MCA, where the MA modules have different formulations of $\Omega_i$ and $\boldsymbol{A}_{mik}$. In deformable DETR, $\Omega_{i}$ is a small set of fractional positions and it cover points on multiple feature maps with different resolutions, and $\Omega_{i}$ is obtained via linear projection over the query $\boldsymbol{q}_i$. However, in DETR $\Omega_{i}$ is all points on the feature map with the lowest resolution. The former way is more efficient and hence supports high resolutions. Moreover, $\boldsymbol{A}_{mik}$ of deformable DETR is also obtained via linear projection, and has a lower computaion cost than DETR, which projects $\boldsymbol{q}_i$ and $\boldsymbol{z}_k$ into new representations and calculates the inner product between them.

\subsection{Set Prediction}
In DETR, the query set $\boldsymbol{q}^t$ at each decoder layer is projected into a bounding box set $b^t$, via two MLPs separately for classification and box regression. Each box set $b^t$ is supervised via the bipartie matching between $b^t$ and GTs. For the bipartie matching, both DETR and Deformable DETR use standard KM algorithm.

\section{Method}
\label{sec:method}

\subsection{Why DETR Fails in Pedestrian Detection?}
We compare both original DETR and deformable DETR on CrowdHuman, a challenging dataset containing highly overlapped and crowd pedestrians, and compare them with Faster-RCNN, a standard baseline in pedestrian detection. All the detector are implemented following the standard hyper-parameters. A counter-intuitive result is illustrated in Table \ref{tab:fasterrcnnvsdetr}. Both original DETR and deformable DETR have obvious performance drop compared with their baselines. Moreover, the training efficiency of DETR on CrowdHuman is much lower than on the common object detection dataset, COCO.

As deformable DETR shows higher performance than original DETR, we take it as our starting point and analyse the reasons behind its performance drop. Our investigation focuses on the decoder, which is the key architecture of DETR's set prediction mechanism. We define $\Omega_i^{c,t}$ as the cross attention field of the $i$-th query on the feature map and $\Omega_i^{s,t}$ as the positions set the $i$-th query makes self attention on in the $t$-th decoder layer.
In the decoder, the query set $\boldsymbol{q}^0$, $\boldsymbol{q}^1$, ..., $\boldsymbol{q}^T$ is iteratively updated. In each decoder layer $t$, $\boldsymbol{q}^{t-1}$ exchange information among each other via self attention over $\Omega^{s,t-1}$, then predict the attention field $\Omega^{c,t-1}$ to extract information of objects from feature maps. At last layer $T$, each query $\boldsymbol{q}_i^T$ matches a GT or background with its attention trajectory $\Omega_i^{c,0}$, $\Omega_i^{c,1}$, ..., $\Omega_i^{c,T-1}$.

\subsection{Dense Queries}
\label{subsec:dq}
\begin{figure*}
	\begin{center}
		
		\includegraphics[scale=0.25]{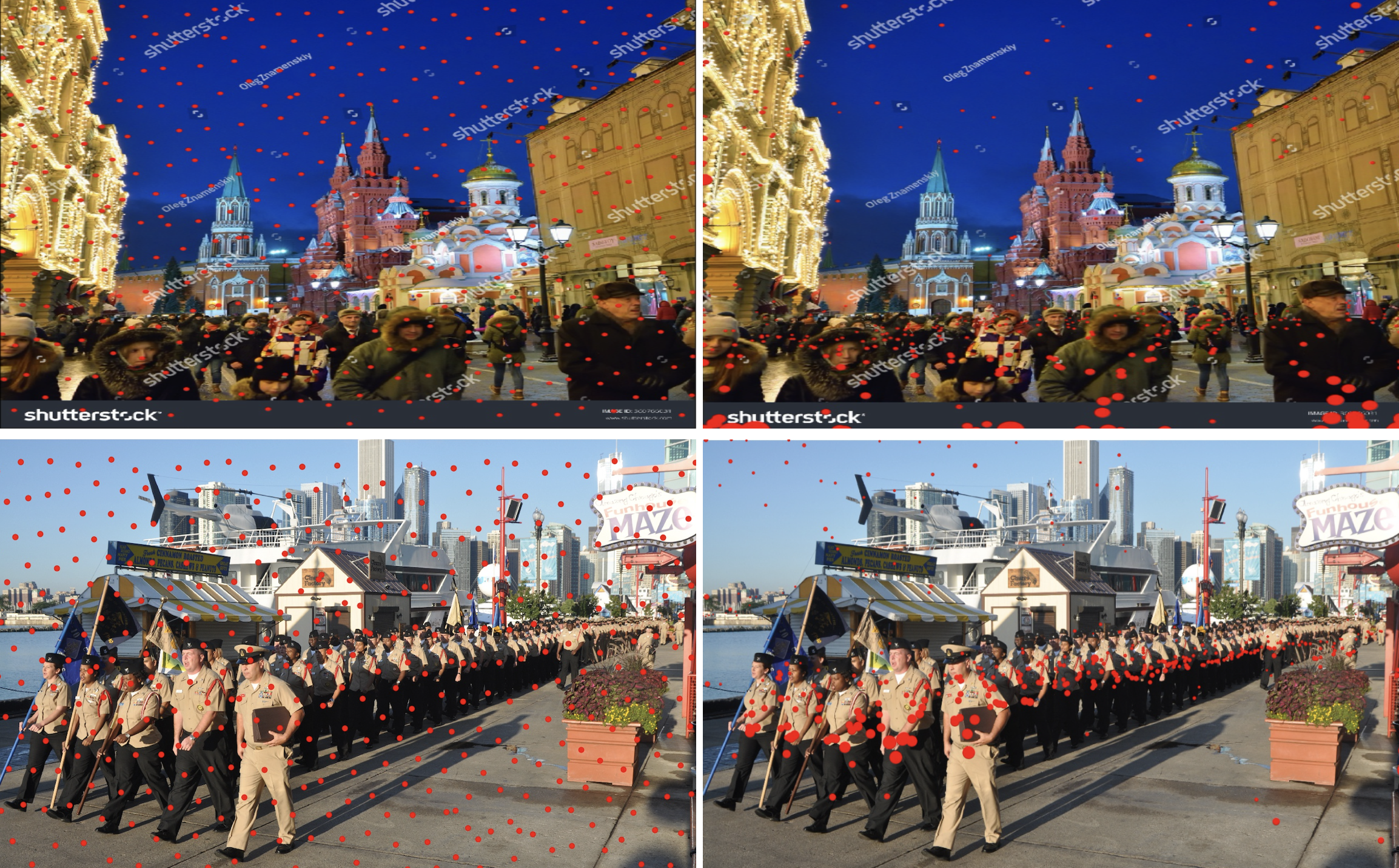}%
		
	\end{center}
	\caption{Visual comparisons between the central positions of the attention fileds $\Omega^{c}$ of the queries in the first decoder layer(left) and the last decoder layer(right). We can observe that object queries need to  learn a shrinking from sparsely uniform distribution to a dense cluster of pedestrian. The size of each circle is representative of the predicted area of each box predicted by the corresponding query.}
	\label{fig:DQfig}
\end{figure*}

\begin{figure*}
	\begin{center}
		
		\includegraphics[scale=0.22]{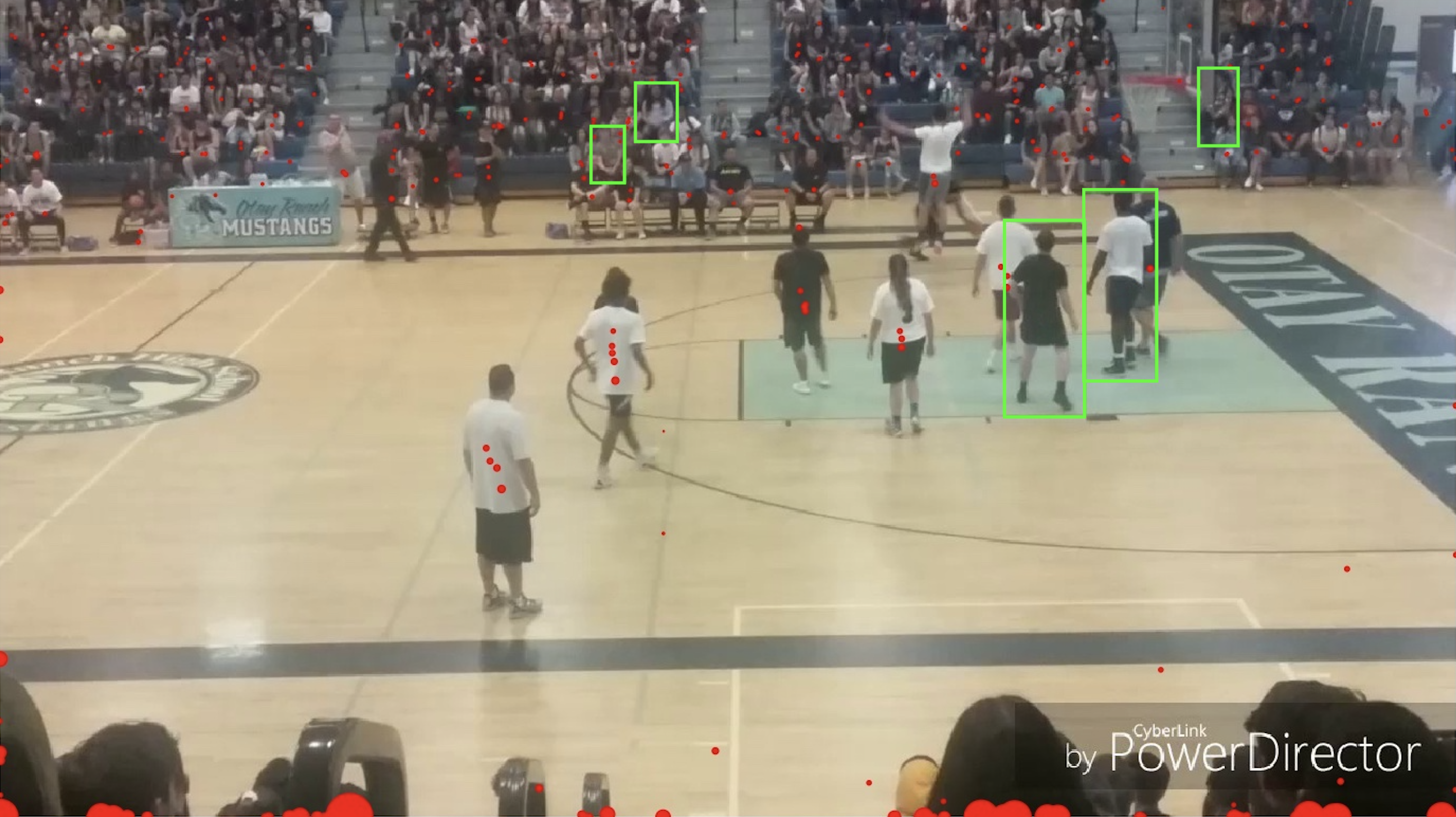}%
		
	\end{center}
	\caption{Prediction of the baseline deformable DETR. We highlight in green GTs that are missed while many other GTs contain more than 3 queries.}
	\label{fig:missedDQfig}
\end{figure*}

First, we find a conflict between the locally \textbf{dense} distribution of GTs and uniformly \textbf{sparse} distribution of query set $\boldsymbol{q}$. Specifically, after training, the initial cross attention fields $\Omega^{c,0}$ of different queries are uniformly distributed on the feature map as shown in Figure \ref{fig:DQfig}(left), due to the uniform appearance of pedestrians. The queries are also sparse due to the limit of computation resources. However, in a single image, pedestrians tend to be distributed densely in some local regions(e.g. a corner or a horizontal line) naturally, while the number of queries $\boldsymbol{q}^0$ whose attention field $\Omega^{c,0}$ locates initially in such local dense regions is not always enough to match all pedestrians lying in them, as in Figure \ref{fig:DQfig}. It means, the decoder layers $F^t$ learns to shrink the attention fields $\Omega^{c,t-1}$ of the uniformly distributed sparse queries progressively from the whole image to compact dense object clusters. This process has two  requirements: (1) a vast perception range of objects and (2) a mapping from a vast range of initial positions to local dense GTs. Two conflicts stands in the two requirements. On one hand, the vast perception range suggest strict requirement on the CNN's reception field. On the other hand, the mapping is highly ambiguous because there are few prior geometry cues for queries to decide how GTs are assigned among them, indeed, although queries are supervised by strict bipartie matching, there is still imbalance between the number of queries lying on different GTs in the dense object clusters. Some GTs are missed while some others contain more than 3 queries, as observed in Figure \ref{fig:missedDQfig}.

The discussions above imply that dense queries will help. When queries are dense enough, the bipartite matching result of each GT is roughly its nearest unique query and the requirement of the query's reception field is much lower. However, the time complexity of the MSA module is quadratic of the query number and hardly bears a dense query setting. We design a \textbf{D}ense \textbf{Q}uery(DQ) module to support a dense setting via reducing the complexity of MSA from $O(N_q^2)$ into $O(N_q)$ of the query number $N_q$.

Local self attention is developed in transformer as an effective way to improve computation efficiency. However, in transformer, queries have one-to-one correspondence to the token positions, and locality can be naturally designed by restricting $\Omega_i^s$ as some near positions $\{...,i-1,i,i+1,...\}$. However, though queries in DETR is equipped with attention positions on feature map, the positions are fractional and variable during training. To develop a distance measure for queries in DETR, we first review what queries should be in $\Omega_i^{s}$ for a certain query $\boldsymbol{q}_i$. In MSA, $\boldsymbol{q}_i$ receives information from queries in $\Omega_i^{s}$ to determine whether itself or another query in $\Omega_i^{s}$ is matched to a GT, as discussed in DETR. This reasonable assumption suggests that the distance should be measured by the possibility that two queries will match the same GT. Consider that each decoder layer predicts its box set $\boldsymbol{b}_i^{t}$ sequentially, we use the overlaps between $\boldsymbol{q}_i^{t-1}$'s box prediction $\boldsymbol{b}_i^{t-1}$ and $\boldsymbol{q}_j^{t-1}$'s $\boldsymbol{b}_j^{t-1}$ in the former decoder layer, to measure the distance of  $\boldsymbol{q}_i^{t-1}$ and $\boldsymbol{q}_j^{t-1}$, because the higher the overlap the more possible that $\boldsymbol{q}_i^{t-1}$ and $\boldsymbol{q}_j^{t-1}$ predict the same GT. Hence we define the distance measure $d_{ij}^{t-1}$ and $\Omega_i^{s,t-1}$ as:
\begin{align}
d_{ij}^{t-1} &= 1-GIOU(\boldsymbol{b}_i^{t-1},\boldsymbol{b}_j^{t-1}),
\label{eq:distance}\\
\Omega_i^{s,t-1} &= \{\boldsymbol{\tau}_{i1}^{t-1},\boldsymbol{\tau}_{i1}^{t-1},...,\boldsymbol{\tau}_{iK}^{t-1}\},
\label{eq:somega}
\end{align}
where $\boldsymbol{\tau}_i^{t-1}$ is the ascending order of $\boldsymbol{d}_{i}^{t-1}$ and we select the nearest $K$ neighbors of $\boldsymbol{q}_i^{t-1}$ based on the defined $\boldsymbol{d}_{i}^{t-1}$.
As shown in Section \ref{experiment}, our DQ algorithm supports twice more queries without calculation cost increase, and achieves even better performance than simply adding twice more queries without changing $\Omega_i^{s}$ via forcing queries to focus only on nearby queries. 


\subsection{Rectified Attention Field}
\label{subsec:rectified}

\begin{figure*}
	\begin{center}
		
		\includegraphics[scale=0.35]{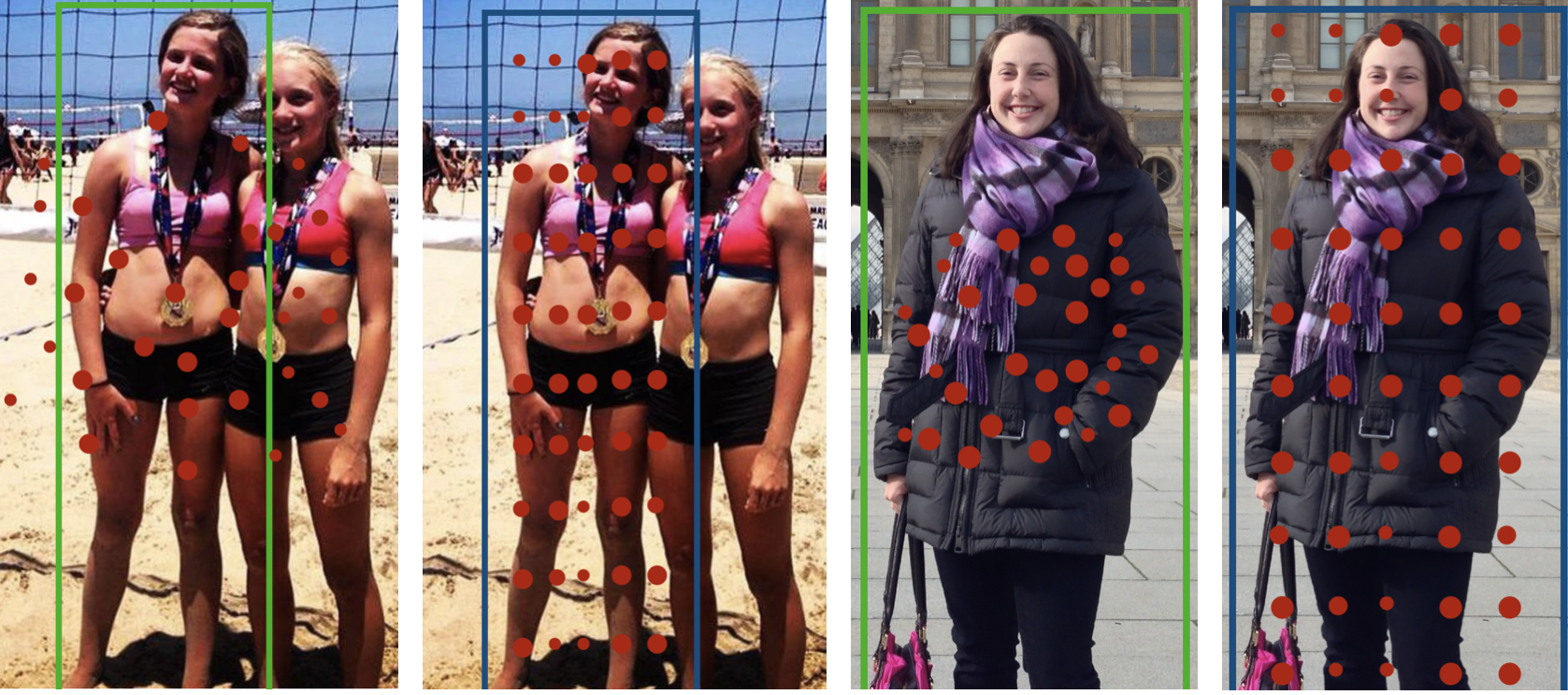}%
		
	\end{center}
	\caption{Visual comparisons between the baseline and the Rectified Attention. The size of each points represent the weights of each sampled point. Pictures with green boxes show attention predicted by deformable DETR for $k=4$.}
	\label{fig:attentionfig}
\end{figure*}

Another problem arises in DETR is that the cross attention $\Omega_i^{c,T}$. $\Omega_i^{c,t}$ is predicted via linear projection over the query feature, it bears a risk to be messy or narrow. In our experiment, in average 34.9\% attention positions in $\Omega_i^{c,t}$ is out of the box of the GT its corresponding query $\boldsymbol{q}_i^t$ matched and 69.7\% of them lies on another nearby GT's box as in Figure \ref{fig:attentionfig}(left), which introduces noise. Furthermore, the learned $\Omega_i^{c,t}$ is often not wide enough for the queries which match large peoples, as in Figure \ref{fig:attentionfig}(right), it harms the accuracy of both the classification score and box regression. 

To relieve the noisy or narrow attention field of the queries, we design a RF(\textbf{R}ectified attention \textbf{F}ield) module to rectify the attention field $\Omega_i^{c,t}$ of the final $M$ layers. As observed in Table \ref{tab:matchingratio}, we find that more than 95\% queries match the same GTs at the last three layers. It means the box prediction $\boldsymbol{b}_i^{t}$ at the 4-th or 5-th layer is nearly always around its final target GT, and we can use the intermediate box prediction $\boldsymbol{b}_i^{t}$ to get a more compact while wide enough attention field $\Omega_i^{c,t}$. We set the attention field $\Omega_i^{c,t}$ as:
\begin{equation}
\begin{aligned}
\Omega_i^{c,t-1}  &= \{(x_i^{t-1}+\frac{i1}{R+1}w_i^{t-1},y_i^{t-1}+\frac{i2}{R+1}h_i^{t-1})\}, \\
& i1, i2 \in \{1,...,R\} ,
\end{aligned}
\label{eq:comega}
\end{equation}
where $x_i^{t-1}$, $y_i^{t-1}$, $w_i^{t-1}$, $h_i^{t-1}$ are position, width and height of the box prediction $\boldsymbol{b}_i^{t-1}$. We use uniform distibuted $R \times R$ points among $\boldsymbol{b}_i^{t-1}$, it relieves the risk of a learned $\Omega_i^{c,t-1}$ to be messy or narrow. Table \ref{tab:RA} shows that our proposed method improves significantly over the baseline on CrowdHuman. The proposed decoder, namely DQRF, is shown in Figure \ref{fig:dqr_architecture}.

\begin{table}[!htbp]
	
	\begin{center}
		\begin{tabular}{l|c|c|c|c|c}
			\hline
			& 2 & 3 & 4 & 5 & 6\\
			\hline
			Similarity   & 0.939 & 0.948 &  0.957 & 0.964 &  0.970 \\
			IoU   & 0.513 & 0.629&  0.754 & 0.841 & 0.908 \\
			\hline
		\end{tabular}
	\end{center}
	\caption{Similarity ratio of matched ground truth between each layer and their previous layer and IoU overlap between each predicted box with their previous layer. Values are computed via averaging over the train set with a pre-trained Deformable DETR on CrowdHuman. }
	\label{tab:matchingratio}
\end{table}

\subsection{Further improvement with visible region}
\label{subsec:fbox}

Recent works \cite{huang2020nms,chi2020relational} develop methods to leverage visible region annotations. These methods reveal that, with marginal extra cost, rational utilization of visible annotations leads to considerable gains. As such, we also propose to leverage visible annotations under the end-to-end framework. The proposed method, \textbf{V-match}, achieves similar performance gain and introduces no extra computational cost.

We design a novel adaptation of the targets of DETR. Considering DETR predicts a sequential sets of boxes $\boldsymbol{b}_1,...,\boldsymbol{b}_T$, we assign the supervision of full boxes on the last $L$ layers while visible boxes on the first $T-L$ layers. It means in the first $T-L$ layers the regression heads predicts visible boxes and the queries are explicitly constrained to focus on the visible part of pedestrians. As shown in Section \ref{subsection:ablation}, V-match achieves stable improvements at zero cost.


\subsection{Other Adaptations}
\label{subsec:otheradaptations}

The training of DETR on dataset with crowd pedestrians has low GPU utilization, because the standard KM algorithm for bipartite matching is cubic of the number of GT. We use the prior that a GT is assigned to the predicted box among its nearest predicted boxes and adapt it accordingly, the resulted Fast-KM has up to 10 times speed up compared to standard KM on CrowdHuman. We omit the details as the complexity of KM algorithm and release it in our code.

We keep the crop augmentation used in DETR and deformable DETR. While it helps little for traditional detectors, it provides diversity of the objects' distribution to prevent DETR from over fitting the objects' distribution in the training dataset. Nevertheless, it is harmful in crowd pedestrian detection since the occluded pedestrian tends to have a small visible part, which is easy to be cropped out. Thus, we adapt the crop operator to conserve at least 80\% area of the visible part of each pedestrian. 

\section{Experiments}
\label{experiment}

\textbf{Datasets}. We evaluate our PED on two human detection benchmarks: CrowdHuman and CityPersons. These two datasets both contains these two categories of bounding boxes annotations: human visible-region and human full-body bounding boxes. As shown in table \ref{tab:datastats} Compared to CityPersons, The CrowdHuman dataset is much more challenging as it contains more instances per images, and those instances are often highly overlapped.

\begin{table}[!htbp]
	\begin{center}
		\begin{tabular}{l|c|c}
			\hline
			Dataset & \#person/img & \#overlaps/img \\
			\hline
			CrowdHuman  & 22.64 & 2.40\\
			CityPersons  & 6.47 & 0.32\\
			\hline
		\end{tabular}
	\end{center}
	\caption{Statistics of each dataset. The threshold for overlap is IoU $>$ 0.5 }
	\label{tab:datastats}
\end{table}

\textbf{Evaluation Metrics}. We evaluate the performance of our PED using two standard metrics used for Pedestrian detection, \eg:
\begin{itemize} 
	\item Average Precision following the standard COCO evaluation metric, which computes the area under the curve of the interpolated precision w.r.t recall curve. This metric is the most commonly used metric in metric in detection as it reflects both the precision and the recall.
	\item Log-average miss rate (MR$^{-2}$), which computes on a log-scale the miss rate on false positive per image with a range of [10$^{-2}$, 10$^{0}$] . This metric is the most commonly used metric in pedestrian detection as it reflects the amount of pedestrian that are not detected.
\end{itemize}

\textbf{Detailed Settings}. Due to the required extra-long training schedule of DETR, we chose to experiment our proposed methods based on Deformable DETR with Iterative Bounding Box Refinement. For all ablation studies, we use as baseline the standard Deformable DETR with Iterative Bounding Box Refinement, all hyper-parameter settings follow Deformable DETR except that we increase the number of queries from 300 to 400, and set the number of queries to 1000 for experiments on our Dense Query method mentioned in Section \ref{subsec:dq}. For training, we use the same protocol as in Deformable DETR, \eg, models are trained for 50 epochs with a learning rate drop by a factor of 10 at the 40$^{th}$ epoch. We also slightly modify the original cropping such that full-body bounding-box are not cropped, since they frequently exceed the size of the images. For our final result, since DETR benefits from longer training schedule we also propose to train the model for 100 epoch with a learning rate drop by a factor of 10 at the  90$^{th}$ epoch. We use our DQ setting, set L to 2 for full box supervision and use 3 RF layers.

\subsection{Ablation Study}
\label{subsection:ablation}

We perform ablation studies and report highest accuracy during training for our new proposed methods in Section \ref{sec:method} on the CrowdHuman dataset. For ablation studies, we replace layers of the deformable DETR with iterative bounding box refinement with our proposed methods starting from the last layer as it is supposed to be the most accurate bounding box prediction. As shown in Table \ref{tab:matchingratio}, at the latter layers, the predicted bounding boxes are less likely to fluctuate. We hypothesize  that replacing first layers does not improve further as table \ref{tab:matchingratio} shows that bounding boxes at the first layers are very noisy and the variation (IoU is low) is high.

\textbf{Ablation study on Rectified Attention Field}. As discussed in \ref{subsec:rectified} the deformable mechanism might yield noisy attention positions among more than one persons and can make the instance inside a detected box ambiguous, hence we propose to add our Rectified Attention Field to adapt decoder layers. Table \ref{tab:RA} shows ablations of our new introduced method by varying the number of decoder layers with our added Rectified Attention Field, our method improves significantly on the baseline under all settings of RF. 





\begin{table}[!htbp]
	
	\centering
	\begin{tabular}{c  c c c c c c  }
		\hline
		\#RF layer    & 0 & 1 & 2 & 3  & 5   \\
		\hline
		
		AP  & 86.74 & 87.70 & 87.87 & \textbf{88.20}  & 87.76  \\
		MR$^{-2}$ & 53.98 & 49.62 & 48.05 & \textbf{46.99}  & 47.38 \\
		
		\hline
	\end{tabular}
	\caption{Effect of the number of Rectified Attention Field(RF) layers. Note that the first row is equivalent to the baseline}
	
	\label{tab:RA}
\end{table}

\textbf{Ablation study on V-Match}. Table \ref{tab:vmatch} shows ablations for the proposed V-match in Sec \ref{subsec:fbox} for different values of L. We can observe that for any setup introducing supervision with visible box improves on the baseline.





\begin{table}[!htbp]
	
	\centering
	\begin{tabular}{c  c c c c c c  }
		\hline
		L      & 6& 5 & 4 & 3 & 2 & 1  \\
		\hline
		
		AP      & 86.74 & 87.47 & 87.38 & 87.35 &\textbf{87.67}& 87.49 \\
		MR$^{-2}$     & 53.98 & 49.78 & 49.25 & 48.94& \textbf{48.07}  & 48.54  \\
		
		\hline
	\end{tabular}
	
	\caption{Effect of the number of layers with full box supervision. Note that in the case L=6, it is equivalent to the baseline since our baseline has 6 decoder layers}
	\label{tab:vmatch}
	
\end{table}

\textbf{Ablation study on Dense Queries}. Table \ref{tab:dq} shows ablation for the Dense Queries algorithm. As mentioned, for fair comparison, we train another baseline Deformable DETR with iterative bounding box refinement and increase the number of queries from 400 to 1000. Our baseline with 1000 queries significantly improves on the baseline with 400 queries, supporting our assumption that a dense query approach is needed. We vary the number of layers with DQ to reduce the computational cost, and observe that applying 5 DQ layers on the baseline with 1000 queries even improves further as it forces queries to only attention on nearby queries rather than irrelevant ones or background. $K$ is fixed to 100 after cross validation.





\begin{table}[!htbp]
	
	\centering
	\begin{tabular}{c c c c c c c c  }
		\hline
		\# DQ layers &   0(no DQ) & 0& 1 & 3  & 5   \\
		\hline
		
		AP  &86.74& 88.34 & 88.81 & \textbf{89.02}  & 88.93  \\
		MR$^{-2}$& 53.98& 49.38 & 48.37 &  \textbf{47.29}    & 47.54 \\
		
		\hline
	\end{tabular}
	
	\caption{Effect of the number of layers with our Dense Queries method. Note the first column is equivalent to the baseline with 400 queries and the latter columns are applying different number of layers of DQ on the 1000 queries baseline.}
	\label{tab:dq}
	
\end{table}

\textbf{Ablation studies on Crop Augmentation}. As discussed in \ref{subsec:otheradaptations} we keep at least 80\% of the visible area of each pedestrian as occluded pedestrian displays only a small visible part, which is easily cropped out Table \ref{tab:crop} shows ablations for our new proposed data augmentation.

\begin{table}[!htbp]
	
	\centering
	\begin{tabular}{c  c c}
		\hline
		Methods & AP & MR$^{-2}$  \\
		\hline
		Baseline       &  86.74  & 53.98   \\
		with Crop Augmentation            &  \textbf{87.17}  & \textbf{51.95}   \\

		\hline
	\end{tabular}
	
	\caption{Ablation study on our Crop Augmentation.}
	\label{tab:crop}
\end{table}

\subsection{Results on CityPersons and CrowdHuman}

CityPersons contains 2,975 and 500 images for training and validation respectively, and CrowdHuman contains 15,000 and 4370 images for training and validation respectively.

\textbf{Comparisons with the standard Faster-RCNN+FPN baseline along with other state-of-the-art methods on CrowdHuman \& CityPersons}. We report results on the Heavy occluded subsets of CityPersons while on CrowdHuman we report AP and MR$^{-2}$. Table \ref{tab:resultcrowd} and \ref{tab:resultcity} show our final results on CrowdHuman and CityPersons. Our proposed PED improves by a large a margin on pedestrian Detection compared with deformable DETR and Faster R-CNN, and is even able to compete with very competitive state-of-the-art methods such as PBM, while having comparable FLOPs.




\begin{table}[!htbp]

	\centering
	\begin{tabular}{c  c c c }
		\hline
		Method & OR-CNN~\cite{zhang2018occlusion} & TLL~\cite{song2018small} & RepLoss~\cite{wang2018repulsion}  \\
		\hline 
		
		Heavy  &55.7 &53.6 &56.9 \\
		\hline 
		Methods         & ALFNet~\cite{liu2018learning} & CSP~\cite{liu2019high} & Ours \\
		\hline 
		Heavy & 51.9& 49.3& \textbf{47.70}\\
		
	\end{tabular}
	
	\caption{Results on Heavy subset on CityPerson shows that our method is effective in occluded and crowded scene.}
	\label{tab:resultcrowd}
\end{table}

\begin{table}[!htbp]

	\centering
	\begin{tabular}{c  c c c}
		\hline
		Method & AP & MR$^{-2}$ & Recall \\
		
		\hline 
		PBM ~\cite{huang2020nms} &  89.3  & \textbf{43.3} & 93.33\\
		Ours &  \textbf{90.08}  & 44.37  & \textbf{93.95}\\
		
		\hline
		Faster-RCNN$^{*}$~\cite{ren2015faster}     & 85.0   & 50.4 & 90.24\\
		AdaptiveNMS$^{*}$ ~\cite{liu2019adaptive}   &  84.7  &  49.7 & 91.27 \\
		Deformable DETR$^{*}$ ~\cite{zhu2020deformable}  & 86.74   & 53.98  & 92.51 \\
		Ours$^{*}$         &  \textbf{89.54}  & \textbf{45.57} & \textbf{94.00} \\
		
		\hline
	\end{tabular}
	
	\caption{Results on CrowdHuman. * stands for no usage of visible boxes}
	\label{tab:resultcity}
\end{table}

\section{Conclusion}


In this paper, we design a new decoder, DQRF, which can be easily implemented and helps alleviate the identified drawbacks of DETR on pedestrian detecion. We also propose a faster bipartite matching algorithm and leverage visible box annotations specifically for DETR to get further improvements. We hope that the resulted detector, namely PED, inspires future work and serves as a baseline for end-to-end pedestrian detection.

{\small
\bibliographystyle{ieee_fullname}
\bibliography{egbib}
}

\end{document}